\documentclass[letterpaper, 10 pt, conference]{ieeeconf}  

\IEEEoverridecommandlockouts                              

\usepackage{amsmath, amsfonts, amssymb} 
\usepackage{gensymb}
\usepackage{graphicx, import} 
\usepackage{caption}
\usepackage{subcaption}
\usepackage{acro}
\DeclareAcronym{NCCR}{
  short = NCCR,
  long  = National Center of Competence in Research,
  short-indefinite = an,
  long-indefinite = a
}

\DeclareAcronym{SLF}{
  short = SLF,
  long  = WSL-Institute for Snow and Avalanche Research,
  short-indefinite = an
}

\DeclareAcronym{sUAS}{
  short = sUAS,
  long  = small uncrewed aerial system,
  short-indefinite = an,
  long-indefinite = a
}

\DeclareAcronym{FMU}{
  short = FMU,
  long  = flight management unit,
  short-indefinite = a,
  long-indefinite = a
}

\DeclareAcronym{GNSS}{
  short = GNSS,
  long  = global navigation satellite system,
  short-indefinite = a,
  long-indefinite = a
}

\DeclareAcronym{VTOL}{
  short = VTOL,
  long  = vertical takeoff and landing,
  short-indefinite = v,
  long-indefinite = v
}

\DeclareAcronym{ICS}{
  short = ICS,
  long  = inevitable collision state,
  short-indefinite = an,
  long-indefinite = an
}

\DeclareAcronym{MILP}{
  short = MILP,
  long  = mixed-integer linear programming,
  short-indefinite = a,
  long-indefinite = a
}

\DeclareAcronym{GIS}{
  short = GIS,
  long  = geographic information system,
  short-indefinite = a,
  long-indefinite = a
}

\DeclareAcronym{SITL}{
  short = SITL,
  long  = software-in-the-loop,
  short-indefinite = an,
  long-indefinite = a
}

\DeclareAcronym{RoC}{
  short = RoC,
  long  = rate of climb,
  short-indefinite = an,
  long-indefinite = a
}

\DeclareAcronym{MPC}{
  short = MPC,
  long  = model predictive control,
  short-indefinite = a,
  long-indefinite = a
}

\DeclareAcronym{MVS}{
  short = MVS,
  long  = multiview stereo,
  short-indefinite = a,
  long-indefinite = a
}

\DeclareAcronym{DEM}{
  short = DEM,
  long  = digital elevation map
}

\DeclareAcronym{OMPL}{
  short = OMPL,
  long  = Open Motion Planning Library
}

\DeclareAcronym{ROS}{
  short = ROS,
  long  = Robot Operating System,
  short-indefinite = a
}

\DeclareAcronym{SAR}{
  short = SAR,
  long  = synthetic-arperture radar,
  short-indefinite = a,
  long-indefinite = a
}

\DeclareAcronym{MCTS}{
  short = MCTS,
  long  = monte-carlo tree search,
  short-indefinite = a,
  long-indefinite = a
}

\DeclareAcronym{MDP}{
  short = MDP,
  long  = Markov decision process,
  short-indefinite = a,
  long-indefinite = a
}
\newcommand{\reffig}[1]{Fig.~\ref{#1}}

\newcommand{\refsec}[1]{Section~\ref{#1}}

\newcommand{\refequ}[1]{Eq.~\eqref{#1}}
\newcommand{\CR}{Cram\'er-Rao}

\usepackage{siunitx}
\usepackage{hyperref}
\usepackage{tikz}
\usetikzlibrary{automata, positioning, arrows}
\tikzset{
->, 
node distance=2.5cm, 
every state/.style={thick, fill=gray!10}, 
initial text=$ $, 
}

\usepackage{cite}
\usepackage{siunitx}
\usepackage{float}
\usepackage{bm}

\usepackage{fancyhdr}
\fancypagestyle{withfooter}{
  
  \fancyfoot[C]{\footnotesize Accepted to the IEEE ICRA Workshop on Field Robotics 2024}
}

\title{\LARGE \bf
Autonomous Active Mapping in Steep Alpine Environments\\with Fixed-wing Aerial Vehicles 
}

\author{Jaeyoung Lim$^{1}$, Florian Achermann$^{1}$, Nicholas Lawrance$^{2}$, Roland Siegwart$^{1}$
\thanks{*This work was supported by ETH Research Grant AvalMapper ETH-10 20-1. We would like to thank Yves B\"{u}hler and Elisabeth Hafner at the \ac{SLF} for their expertise and support in avalanche mapping and modeling. We would also like to thank David Rohr$^{1}$ at ETH Zurich, for being a reliable safety pilot for countless flight tests.}
\thanks{$^{1}$ Autonomous Systems Lab, ETH Z\"urich, Z\"urich 8092, Switzerland {\tt \footnotesize \{jalim, acfloria, rsiegwart\}@ethz.ch}}%
\thanks{$^{2}$ CSIRO Robotics, Data61, QLD 4069, Australia, { \tt\footnotesize nicholas.lawrance@csiro.au}}%
}

\begin{document}

\maketitle
\thispagestyle{withfooter}
\pagestyle{withfooter}

\begin{abstract}
Monitoring large scale environments is a crucial task for managing remote alpine environments, especially for hazardous events such as avalanches. One key information for avalanche risk forecast is imagery of released avalanches. As these happen in remote and potentially dangerous locations this data is difficult to obtain. Fixed-wing vehicles, due to their long range and travel speeds are a promising platform to gather aerial imagery to map avalanche activities. However, operating such vehicles in mountainous terrain remains a challenge due to the complex topography, regulations, and uncertain environment. In this work, we present a system that is capable of safely navigating and mapping an avalanche using a fixed-wing aerial system and discuss the challenges arising when executing such a mission. We show in our field experiments that we can effectively navigate in steep terrain environments while maximizing the map quality. We expect our work to enable more autonomous operations of fixed-wing vehicles in alpine environments to maximize the quality of the data gathered.
\end{abstract}



\section{INTRODUCTION}

Environment monitoring is crucial for managing remote alpine environments. For example, snow avalanche is the most hazardous natural hazard in Switzerland~\cite{schweizer_snow_2008}. {SLF} maintains an avalanche bulletin, which keeps track of avalanches and reports avalanche risk levels to the public. Up-to-date and complete avalanche activity maps would enable a direct improvement of the avalanche bulletin, including sound validation of its accuracy and reliability, aiding the effort to combat avalanche-related injuries and deaths. 

However, acquiring high quality avalanche activity data remains difficult to access~\cite{Schweizer2021}. Avalanche data relies on observer reports or static sensor stations which require significant in-situ infrastructure that can be difficult and expensive to install, maintain, and access. Satellite images can be used for monitoring avalanches using visual~\cite{hafner_automated_2022} or \ac{SAR}~\cite{eckerstorfer_near_2019} data, but are expensive and hard to access.

Recently, robotic systems are increasingly being deployed as data-gathering tools for environment monitoring~\cite{dunbabin_robots_2012}. 
In particular, easily manageable \acp{sUAS} have been deployed for various applications~\cite{bircher_three-dimensional_2016, shah_multidrone_2020, islam_real_2021} due to their capability to provide unique viewpoints and reach hard-to-access areas. 
Utilizing \ac{sUAS} in alpine environments for environment monitoring would allow access to high-quality data, without requiring vast infrastructure. Also, the use of autonomous vehicles would open opportunities for active information-gathering techniques~\cite{bajcsy_active_1988}, where observations can be taken adaptively, to maximize the information of the gathered data~\cite{das_data_2015, arora_multi_2019, popovic_informative_2020} and robust against environment uncertainties~\cite{preston_phortex_2024}. 

\begin{figure}[t]
\vspace*{0.2cm}
\centerline{\includegraphics[width=\linewidth]{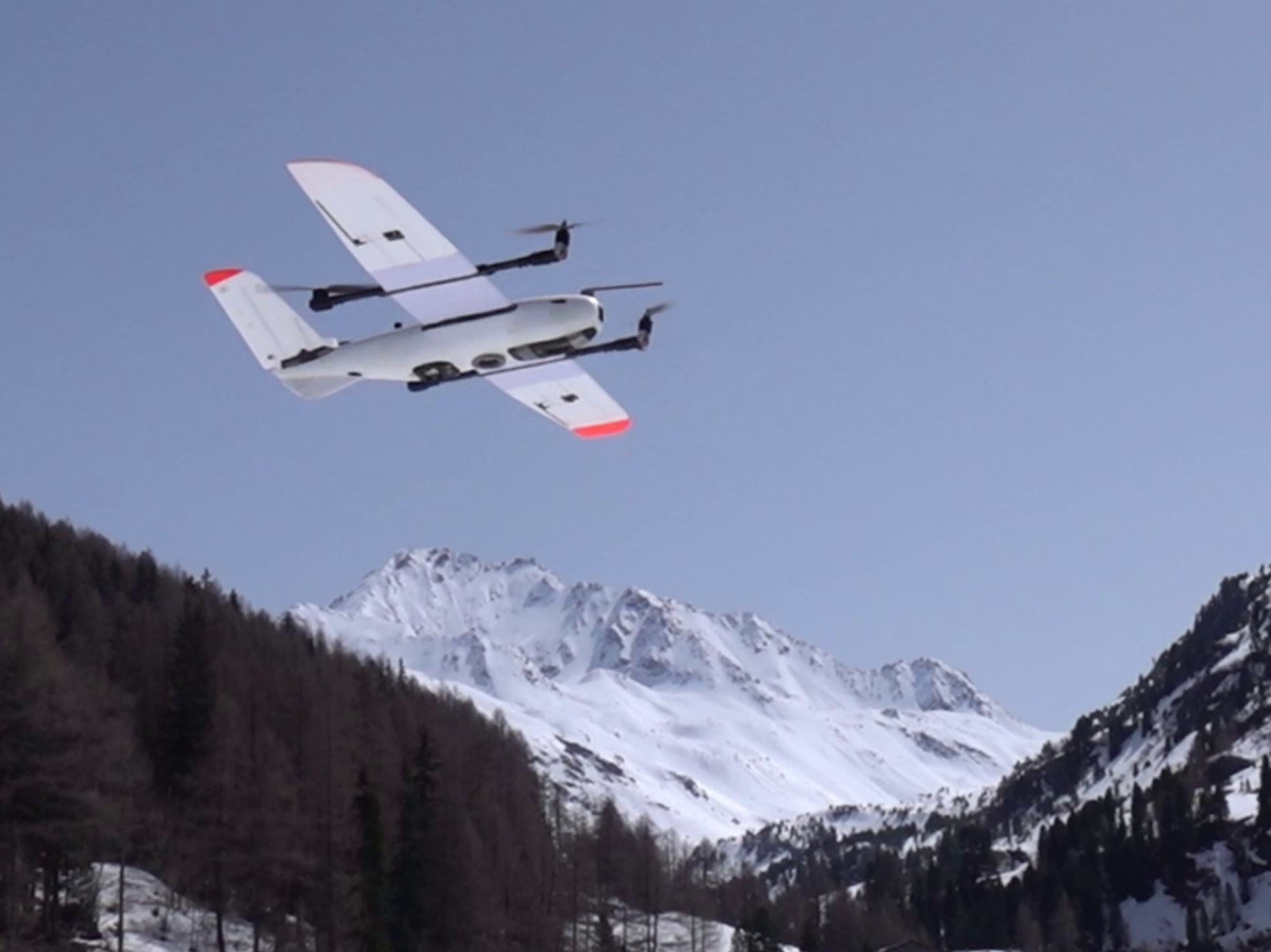}}
\caption{Image of the tiltrotor VTOL platform during take-off during the field test in a narrow valley in Davos, Switzerland.}
\label{fig:paper_overview}
\end{figure}

In particular, fixed-wing type aerial vehicles have become increasingly popular, due to their aerodynamic efficiency, can cover large areas enabling access to remote environments such as glaciers~\cite{oettershagen_robotic_2018, jouvet_high_2019, buhler_photogrammetric_2017, teisberg_development_2022}, hurricanes~\cite{lin_eyewall_2008} and volcanoes~\cite{astuti_overview_2009}. 

However, operating fixed-wing aerial vehicles close to the terrain in steep alpine environments remains a major challenge. Fixed-wing vehicles operate at high speeds and are limited in maneuverability. This increases the risk of the vehicle entering an unsafe state, as the terrain might become steeper than the vehicle can climb, or narrow than the vehicle can turn~\cite{lim_safe_2024}. 

This workshop paper presents a fixed-wing \ac{sUAS} capable of autonomously navigating alpine environments and actively mapping an avalanche. The vehicle traverses the steep alpine environment using a safe path planner~\cite{lim_safe_2024} to navigate to the region of interest for mapping. The system then collects high-quality photogrammetry data by optimizing viewpoints on the fly for reliably creating a photogrammetry reconstruction of an avalanche. We demonstrate and evaluate the approach on an integrated platform on a tiltrotor \ac{VTOL} platform in a field deployment in real alpine terrain in Davos, Switzerland. \reffig{fig:paper_overview} shows a photo taken during a field test.

While this system is built for active avalanche mapping, we expect this work to enable other close-terrain information-gathering applications using fixed-wing \ac{sUAS}.

\section{CORE CHALLENGES}\label{sec:core_challenges}
In this section, we discuss the core challenges towards enabling fixed-wing \ac{sUAS} operations for environment monitoring in steep alpine environments.

\subsection{Safe Navigation in Steep Terrain}
Safely navigating in the target environment is crucial to access regions of interest. Fixed-wing vehicles have limited maneuverability (minimum turning radius and \ac{RoC}) due to the use of aerodynamic lift to stay airborne~\cite{chitsaz_time-optimal_2007}. As terrain in alpine environments can be steeper than the \ac{RoC} or require tighter turns than the vehicle's minimum turn radius, vehicles can encounter~\acp{ICS}. \Iac{ICS} is part of the state space where no feasible input can prevent a future collision with the environment~\cite{fraichard_inevitable_2004}. Correcting a maneuver that would enter~\iac{ICS} can be challenging for a manual operator, as the inevitable collision may be far into the future.

This property has typically limited deployments of fixed-wing vehicles to large open spaces. Operating fixed-wing vehicles near terrain would enable close-up information-gathering tasks traditionally performed by less efficient multi-rotor vehicles.
Furthermore, near-surface operations may become necessary as recent regulations require \acp{sUAS} to stay within \SI{120}{\metre} from the closest point of the terrain surface~\cite{eu2019commission}.

Handling \acp{ICS} has been addressed in various systems, where most approaches use emergency maneuvers to ensure safety~\cite{arora_principled_2014, bekris_avoiding_2010}. However, the inability to hover with fixed-wing vehicles makes the problem more challenging. 

Ensuring safety can be even more challenging in the presence of dynamic obstacles. Fixed-wing vehicles operate in shared airspace, with other aircraft or paragliders, whose behaviors can be hard to detect or predict.

\subsection{Handling Environment Uncertainty}

Flying in alpine terrain is challenging as the environment constantly changes, and the state of the environment cannot be precisely known. While \ac{GIS} provides detailed elevation maps, they are incomplete as they do not contain vegetation, elevation data of structure (e.g. cable cars or power lines), or temporary build objects or changes in the environment. Locally detecting and mapping such obstacles is challenging due to the required detection range necessary to plan and execute an avoidance maneuver~\cite{hinzmann_flexible_2019}. Furthermore, camera and lidar-based sensors lack the resolution for detecting thin objects, such as powerlines.

The uncontrolled airspace close to the ground where \ac{sUAS} typically operates might encounter other air traffic. While manned aircraft equipped with transponders are easy to detect, other aircraft such as paramotors, hang gliders might not be equipped with transponders. As aircraft operating under this airspace fly under visual flight rules~\cite{wilson2018flightrules}, \iac{sUAS} operating in this space will also need to detect and avoid other incoming air traffic. However, detecting other air traffic using visual sensors can be challenging, and the problem is still an ongoing research topic~\cite{vitiello2024senseandavoid}.

In alpine environments, wind can exceed the cruise speed of \ac{sUAS}~\cite{stastny_flying_2019}, and exhibit high spatial gradients close to the terrain~\cite{achermann_windseer_2024}, thus leading to deviations from the preplanned flight path compromising safety. The available weather forecasts only provide wind forecasts at kilometer-scale~\cite{voudouri2018cosmo}, which is not sufficient for planning \ac{sUAS} flight operations. Ongoing research has demonstrated computing local high-resolution wind fields~\cite{achermann_windseer_2024} and planning safe and efficient paths in such wind fields~\cite{oettershagen2017towards, duan_energy_2024}. However, integrating these approaches into a flight stack and evaluation in field experiments still needs to be demonstrated.



\subsection{Online Decision Making for Information Maximization}

Currently, the most common approach for information gathering is done by preplanned missions by experts~\cite{frolov_can_2014}. An example is using boustrophedon decomposition, where a region of interest is decomposed into sweep patterns such that the sensor covers the region of interest~\cite{choset_coverage_2001, bahnemann_revisiting_2019}. However, these approaches assume that the region of interest is a polygon and that vehicle dynamics are holonomic. This makes it challenging to incorporate the nonholonomic dynamic constraints of fixed-wing vehicles. Also, preplanned approaches cannot adapt to disturbances, such as the captured viewpoints differing from the planned ones.

Active perception approaches~\cite{bajcsy_active_1988}, which autonomously adjusts the flight path based on the data acquired during the mission so far, could enable better quality data acquired than expert-planned paths. 

The high-speed and dynamic nature of fixed-wing vehicles requires real-time decision-making. However, most informative planning applications require online reconstruction in the loop~\cite{hepp_plan3d_2019, morilla_sweep_2022, schmid_efficient_2020}. Reconstruction processes such as photogrammetry or \ac{MVS} may even be prohibitive to process online with full image resolutions. 
Therefore, the core challenge is to create a decision-making framework where we can represent information in a compact form such that decisions can be made online in real-time while navigating at high speed.

\section{SAFE NAVIGATION AND OPERATIONS}\label{sec:safe_navigation}

In this section, we address the challenges in safety for navigating in steep alpine environments while staying under the distance constraints posed by regulations.

\subsection{Safe Planning}
The safe planning problem can be formulated as finding a path $\bm{\eta}$ between the start set $\mathcal{X}_{start}$ to goal set $\mathcal{X}_{goal}$ where all states along the path should not be in~\iac{ICS}. However, explicitly evaluating whether a state is~\iac{ICS} is impractical, as it involves infinite horizon collision checks~\cite{fraichard_inevitable_2004}.

\begin{figure}[t]
    \centering{\includegraphics[width=0.8\linewidth]{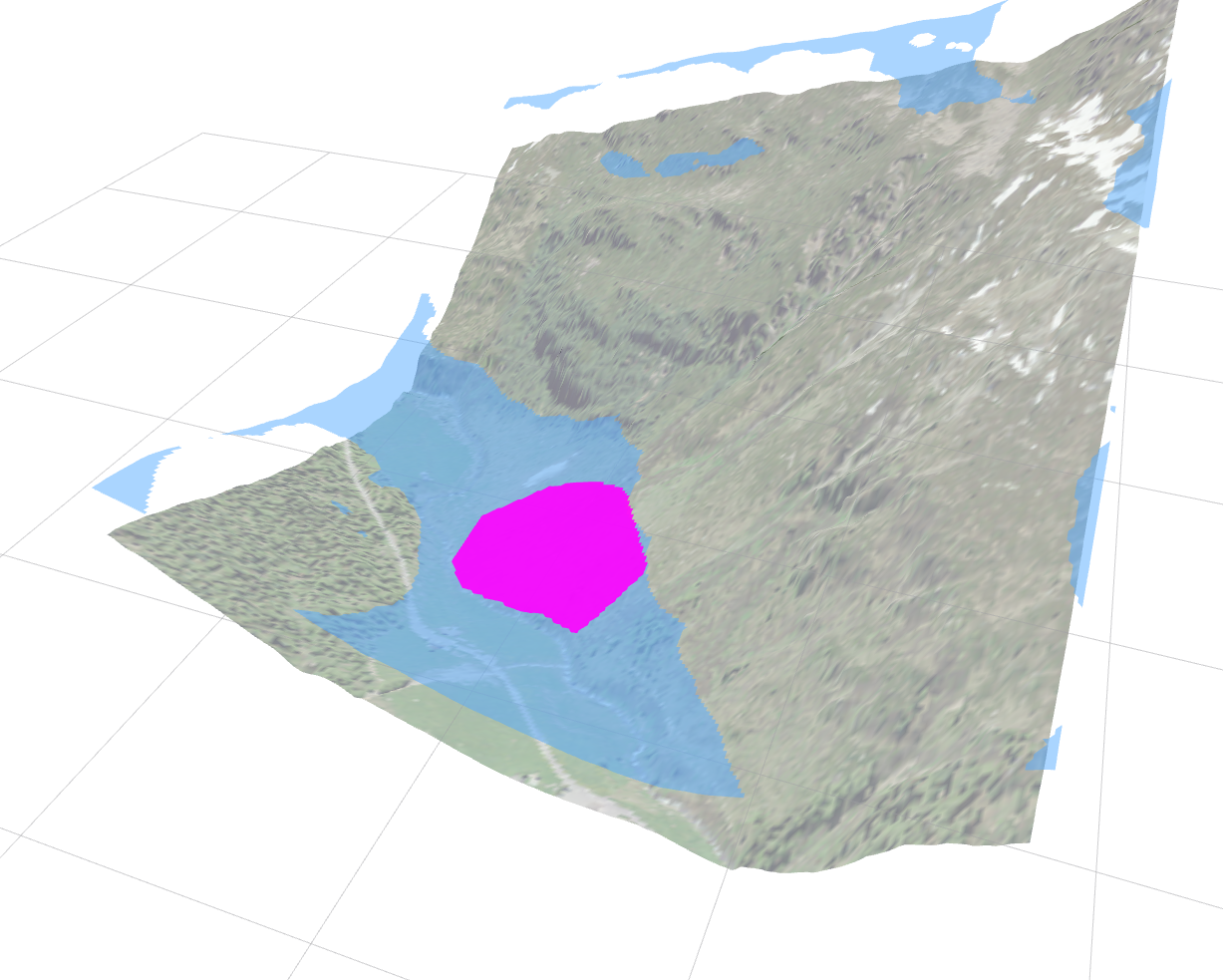}}
    \caption{Valid loiter positions overlaid as a blue surface over the terrain where the field test was conducted. The minimum and maximum distance constraint is defined as \SI{50}{\metre} and \SI{120}{\metre}. The magenta region shows the region of interest.}
    \label{fig:terrain_safetymask}
\end{figure}

We use the approach proposed in~\cite{lim_safe_2024}, where circular loiter paths are used for precomputing safe start and goal states. This effectively abstracts the goal set as a 2-dimensional position, which can be evaluated quickly. The blue overlay in \reffig{fig:terrain_safetymask} shows the valid loiter positions, where the height is defined by the average between the distance constraints~\cite{lim_safe_2024}. Note that the vehicle is still able to traverse through the regions that are not marked as valid loiter regions, but are not able to stop and loiter while complying with the distance constraints.

To utilize the safe navigation planner for operations, we create two different behaviors of planner behavior by modifying how the start and goal sets $\mathcal{X}_{start}, \mathcal{X}_{goal}$ are generated.

\subsubsection{Loiter-to-Loiter Navigation}
This is the nominal mode of operation, where the start and goal sets are both sampled from a circular path. We denote the clockwise loiter path as $P^{CW}(\bm{c}, R)$ and counter-clockwise loiter path $P^{CCW}(\bm{c}, R)$, where $\bm{c}$ is the position of the loiter center and $R$ is the radius. Then the start and goal set can be generated as~\refequ{eq:loiter-to-loiter}.

\begin{align}
    \mathcal{X}_{start} &= P^{CW}(\bm{c}_{start}, R) \;or\; P^{CCW}(\bm{c}_{start}, R)\\
    \mathcal{X}_{goal} &= P^{CW}(\bm{c}_{goal}, R)\cup P^{CCW}(\bm{c}_{goal}, R)\nonumber
    \label{eq:loiter-to-loiter}
\end{align}
Note that the goal set is a union of both directions of the loiter paths, while the start set is constrained on which direction the vehicle is loitering the circular path.

\subsubsection{State-to-Loiter Navigation}\label{sec:emergency_abort}
This mode is similar to the loiter-to-loiter navigation, except that the start set $\mathcal{X}_{start}$ consists of a single state. Such behavior is useful when replanning the path from the end of the currently tracked path to modify the planned path during an emergency abort. 
\begin{align}
    \mathcal{X}_{start} &=\bm{x}_{end}\\
    \mathcal{X}_{goal} &= \bigcup_{i} P^{CW}(\bm{c}_i, R_i)\cup P^{CCW}(\bm{c}_{i}, R_i)\nonumber
    \label{eq:state-to-loiter}
\end{align}
Also, it is possible to specify multiple candidate goal positions by expanding the $\mathcal{X}_{goal}$. Such a feature is useful when multiple goal positions are specified as part of the planning problem.

\subsection{Operation States}
\begin{figure}[t]
\centering 
\begin{tikzpicture}[]
    \node[state, initial, fill=yellow] (q1) {Hold};
    \node[state, above of=q1, fill=green] (q2) {Navigate};
    \node[state, right of=q2, fill=blue, text=white] (q3) {Mapping};
    \node[state, left of=q2, fill=red] (q4) {Abort};
    \node[state, below of=q3, fill=cyan] (q5) {Return};
    \draw (q1) edge[bend left, dashed] node{} (q2)
    (q1) edge[below, dashed] node{} (q3)
    (q2) edge[bend left, below] node{} (q1)
    (q3) edge[bend right, above, dashed] node{} (q4)
    (q2) edge[bend right, above, dashed] node{} (q4)
    (q4) edge[bend right, above] node{} (q1)
    (q1) edge[bend right, above, dashed] node{} (q5)
    (q5) edge[bend right, below] node{} (q1);
  \draw (current bounding box.south west) ++(-0.0,-0.0) coordinate (tmp) -- +(1.0,0) node[right] {Completed};
  \draw [dashed] (tmp)  ++(0,-0.5) -- +(1.0,0) node[right] {Commanded};
\end{tikzpicture}
\caption{Finite state machine of vehicle operations. Dotted transitions are triggered by the operator, and the Solid transitions are triggered upon task completion of the state. Each state is color-coded uniquely for visualizations in~\reffig{fig:davos_reference_altitude}~\reffig{fig:hinwil_mapping_results}}
\label{fig:fsm}
\end{figure}
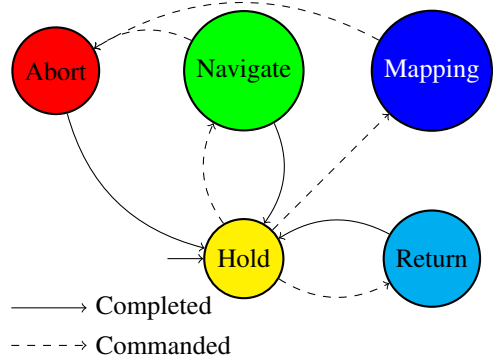

Taking advantage of the planning formulation from the previous section, we operate the vehicle through a finite state machine, where the state machine is visualized in~\reffig{fig:fsm}. Transitions between states are visualized by arrows, where the dotted transitions are triggered by the operator, and the solid transitions triggered automatically by termination conditions. Each state corresponds to a specific task that the vehicle is doing:
\subsubsection{Hold}
In a \emph{Hold} state, the vehicle is following a circular loiter path. Since the circular path is periodic, the vehicle is in a safe state, where it can indefinitely wait for the next operator command. When tasks are successfully finished in \emph{Navigate}, \emph{Return}, \emph{Abort}, the vehicle transitions to the \emph{Hold} state. It is assumed that a mission starts in a \emph{Hold} state. 

\subsubsection{Navigate}
The \emph{Navigate} state is used for navigating between loiter paths. The planner is operated as a \emph{Loiter to Loiter Navigation} as described in~\refequ{eq:loiter-to-loiter}. The vehicle can enter a \emph{Navigate} state when the path planner has found a safe path as described in~\refsec{sec:safe_navigation}. When the vehicle reaches the goal loiter, the state transitions back to \emph{Hold}. The plan can be aborted while in transit by transitioning to the \emph{Abort} state.

\subsubsection{Mapping}
During the \emph{Mapping} state, the active mapping planner is executed as described in~\refsec{sec:active_mapping}. The termination of mapping is triggered by transitioning to \emph{Abort}, which looks for a safe place for the vehicle to loiter.

\subsubsection{Abort}
The \emph{Abort} state is used for aborting a currently executed task. The planner is formulated as in \emph{state-to-loiter}, where the end of the current segment is set as the start state. To identify safe rally points, multiple points within a specified radius are sampled from the end of the current segment. If a sampled position is in the \emph{valid loiter region}, then valid rally points are appended to the goal set $\mathcal{X}_{goal}$. When the vehicle reaches the final loiter, the \emph{Abort} state is transitioned back to \emph{Hold}.

\subsubsection{Return}
\emph{Return} state works similarly to the \emph{Navigate} state, where the start set is defined by the current loiter and the goal set is defined as the loiter at the initial launch position.

\section{Online Active Aerial Photogrammetry}\label{sec:active_mapping}


In this section, integrate an information-gathering strategy for optimizing viewpoints for aerial photogrammetry. \cite{lim_fisher_2023} have shown that a fisher-information-based view utility function can be used for active view planning. However, the motion primitive tree is exhaustively evaluated, therefore the maneuver planning is prohibitive to evaluate in real-time. 

Therefore, we use an anytime approximate graph search algorithm \ac{MCTS}, to evaluate the motion tree in real-time. We first formulate the problem as~\iac{MDP} consisting of a tuple $(\mathcal{S}, \mathcal{A}, \mathcal{P}, \mathcal{R})$.

\subsubsection{State Space $\mathcal{S}$}
The state $\mathcal{S}$ consists of the vehicle state $X$ and the map uncertainty $\bm{I}$. The vehicle state consists of the position and a heading. The map uncertainty $\bm{I}$ consists of the Fisher-information matrix of all the landmark positions.

\subsubsection{Action Set $\mathcal{A}$}
An action $\bm{a}$ is defined by the change of curvature and flight path angle. The action set $\mathcal{A} $consists of three discrete curvature changes and flight path angles, which make 9 motion primitives $\mathcal{A} = \{-\Delta\kappa, 0, \Delta\kappa\}\times\{-\gamma_{max}, 0, \gamma_{max}\}$. if the change of curvature exceeds the maximum curvature $\kappa_{max}$, we generate three different maneuvers with the same curvature changes that satisfy the maximum curvature constraint.

\subsubsection{State Transition $\mathcal{P}$}
The state transition of the vehicle states is defined by the Dubins airplane model~\cite{chitsaz_time-optimal_2007}.
\begin{align}
    \begin{pmatrix}
        \Dot{x}\\
        \Dot{y}\\
        \Dot{z}\\
        \Dot{\theta}
    \end{pmatrix} = V\begin{pmatrix}
        \cos{\gamma}\cos{\theta}\\
        \cos{\gamma}\sin{\theta}\\
        \sin{\gamma}\\
        \kappa
    \end{pmatrix}
\end{align}
The map information update is done by updating the fisher information by the acquired view set as in~\cite{lim_fisher_2023}.

\subsubsection{Reward $\mathcal{R}$}
The view utility metric can be computed as the difference in map quality with the added viewpoints. Therefore, the reward of a rollout can be computed as~\refequ{eq:view_utility}, where $\mathbf{v}'$ is the set of viewpoints acquired by a specific rollout.

\begin{align}
    R(\mathbf{v}' | \mathbf{v}) = Q_{\mathcal{L}}\left(\mathbf{v}\right) - Q_{\mathcal{L}}\left(\mathbf{v} \cup \mathbf{v}' \right).
    \label{eq:view_utility}
\end{align}
We solve this problem with the UCB1 algorithm~\cite{browne2012survey} for \ac{MCTS}, where the action is expanded with the upper confidence bound. We further normalize the reward function based on the map utility to keep the exploration of the behavior constant.

Since each action modifies the uncertainty map, this would require every change in the map to be stored in the tree. To reduce the required memory of the tree, we compute action reward only after the full rollout. This additionally speeds up the rollout since the reward calculation is done in batches.


\begin{figure}[t]
\vspace*{0.2cm}
\centerline{\includegraphics[width=\linewidth]{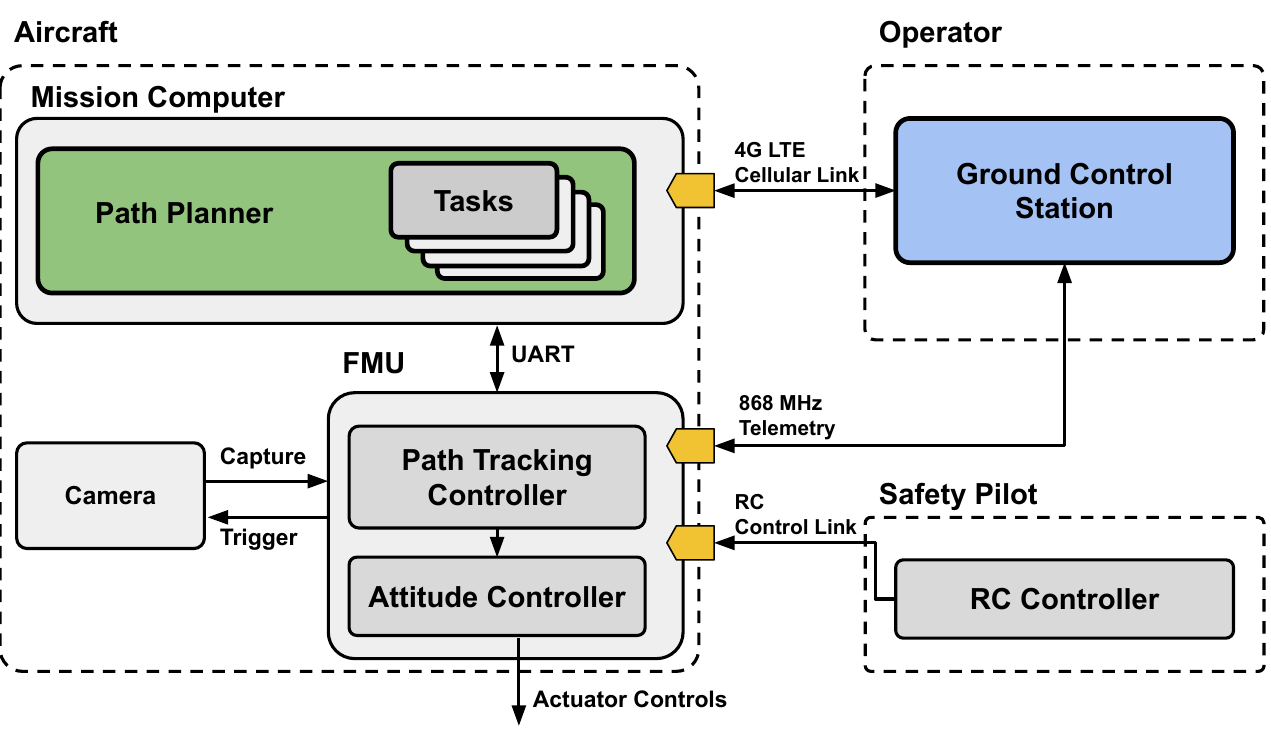}}
\label{fig:communication_flow}
\caption{Overview of the system that was used for flight testing. The system consists of a Mission computer, \iac{FMU} which are controlled by an operator and a safety pilot.}
\label{fig:system_overview}
\end{figure}



\section{VEHICLE PLATFORM}\label{sec:platform}
We present an aerial vehicle capable of safely navigating and actively mapping a region of interest in steep alpine terrain. \reffig{fig:system_overview} shows the overview of the system. 

\subsection{Hardware Setup}
The platform shown in \reffig{fig:paper_overview} is a tiltrotor \ac{VTOL} aircraft with a mass of \SI{5.7}{\kg} and a wingspan of \SI{2300}{\mm} based on the Makefly Easy Freeman. The wing-mounted motors tilt upwards to hover during takeoff and landings. This capability eliminates the need for a flat runway and allows the vehicle to launch and land in steep and confined spaces. The front rotors are tilted forward for the remainder of the mission, such that the vehicle flies efficiently as a fixed-wing vehicle.

For the payload, the system is equipped with a 61 MP Sony A7R mirrorless camera for mapping. The camera capture signal is wired to the \ac{FMU}, to acquire accurate timestamps for geotagging. An RTK GNSS is used for global position estimation. 

The \ac{FMU} runs low-level control loops and runs the PX4 autopilot for \acs{GNSS}-based navigation. The platform uses an Intel NUC, equipped with a \SI{3.5}{\giga\hertz} Intel Core i7-7567U CPU as the onboard computer, which runs the path planners.

The operator communicates to the mission computer through a cellular connection. The operator sends high-level commands over the cellular connection and the telemetry data from the plane is visualized on the \ac{DEM} for planning. An additional \SI{868}{\mega\hertz} telemetry connection is used for redundancy, and an RC link connects a safety pilot to the vehicle to abort the mission in case of an emergency. 

\subsection{Software Setup}
The operator sends the necessary information for operations such as goal positions, and the transition between different tasks. Depending on the mode, the path planner executes different planner tasks to find the reference path. A reference path consists of a geometric path defined by its start position, length and curvature. The mission computer continuously sends path-tracking reference commands $\mathbf{r} = [\mathbf{p}, \mathbf{v}, \kappa]$ from the closest point on the Dubins airplane path $\mathbf{p}$, tangent $\mathbf{t}$, and curvature $\kappa$ to the \ac{FMU} at \SI{10}{\hertz}. The reference is passed to a nonlinear path following a guidance controller based on~\cite{stastny_flying_2019}, which is robust against excess wind.

\section{EVALUATION}
\subsection{Setup}
\begin{figure*}[t]
\vspace*{0.2cm}
\begin{subfigure}{0.3\linewidth}
\centerline{\includegraphics[width=\linewidth]{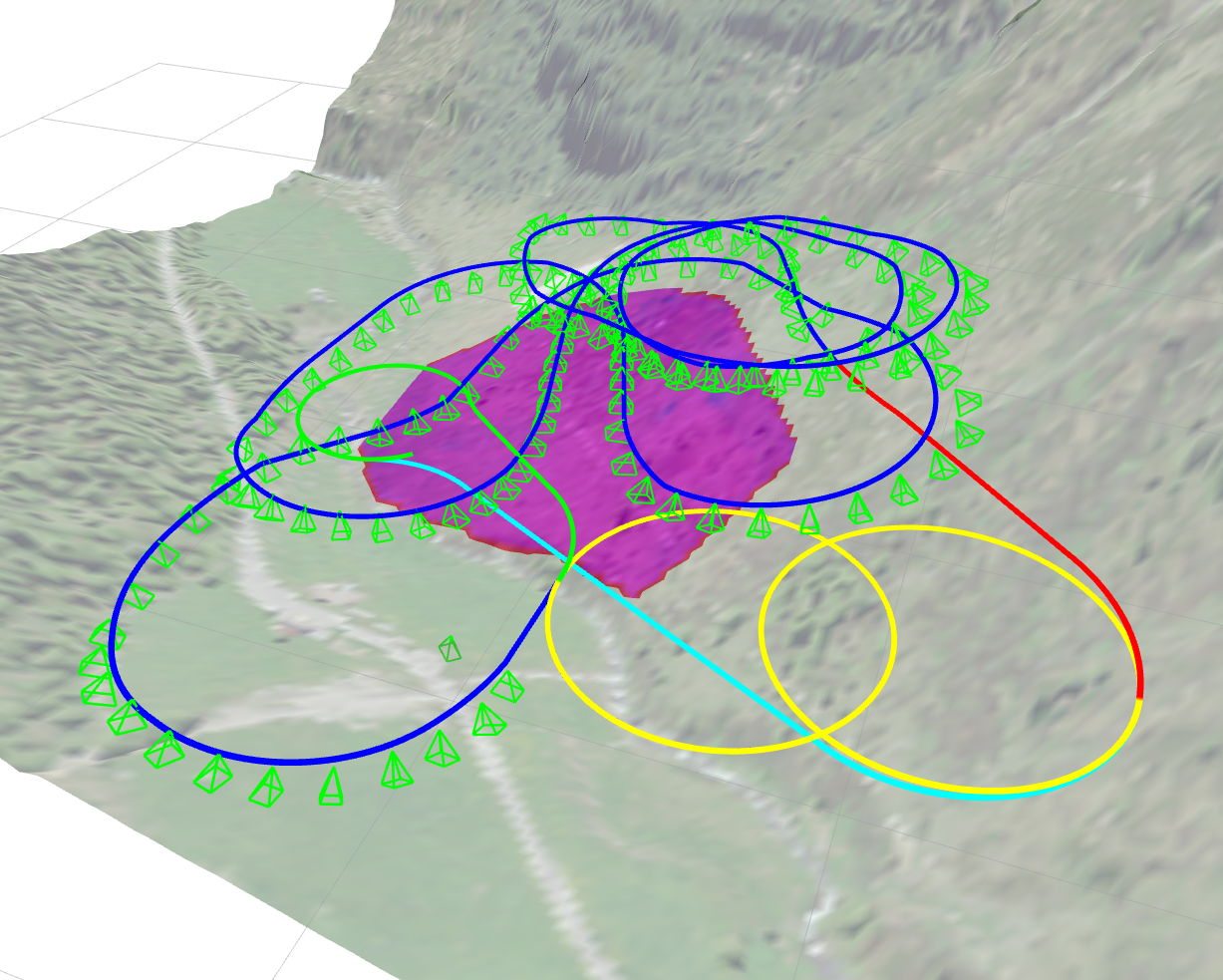}}
\caption{Active Mapping.}
\label{fig:hinwil_mapping_results:a}
\end{subfigure}
\begin{subfigure}{0.3\linewidth}
\centerline{\includegraphics[width=\linewidth]{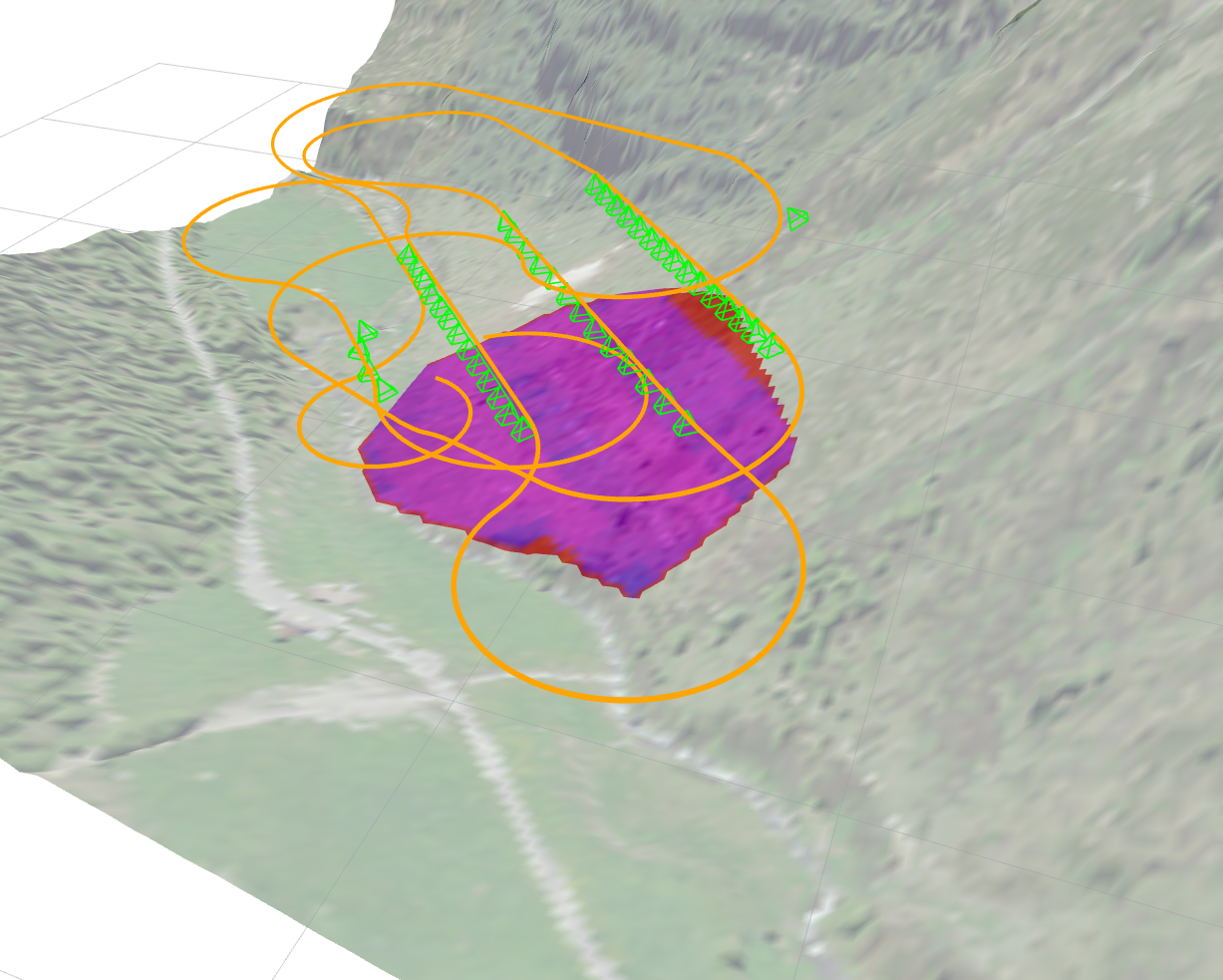}}
\caption{Coverage Mapping.}
\label{fig:hinwil_mapping_results:b}
\end{subfigure}
\begin{subfigure}{0.39\linewidth}
\centerline{\includegraphics[width=\linewidth]{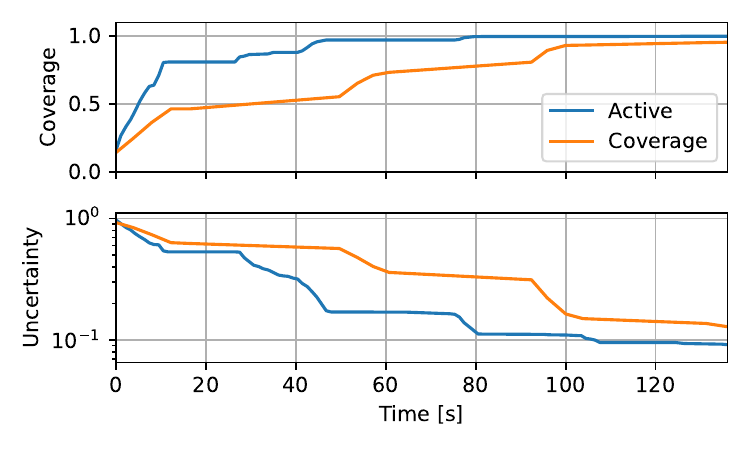}}
\caption{Expected Coverage and Uncertainty.}
\label{fig:hinwil_mapping_results:c}
\end{subfigure}
\caption{Viewpoints generated from (a) coverage mapping (b) active \ac{MCTS}-based mapping in the test area, where each of the view frustums are visualized in green. The target region of interest is shown in magenta.}
\label{fig:hinwil_mapping_results}
\end{figure*}

We evaluate the approach to the described system. The goal is to map a flat polygonal area with the extend of \SI{390}{\metre}$\times$\SI{295}{\metre} located in Davos, Switzerland. The region of interest is overlaid with magenta~\reffig{fig:hinwil_mapping_results}. We compare our active mapping method against a terrain-aware coverage plan generated by sweep patterns, a widely used mission planning tool. The photogrammetry reconstruction was done using Agisoft, and was geo-aligned using the geostags generated by the capture signal from the camera connected to the \ac{FMU}.

\subsection{Safe Navigation}
We deploy our path planner for navigation and mapping. Given that the target environment was flat, we evaluate the tracking performance of the planned path.

\begin{figure}[t]
\centerline{\includegraphics[width=\linewidth]{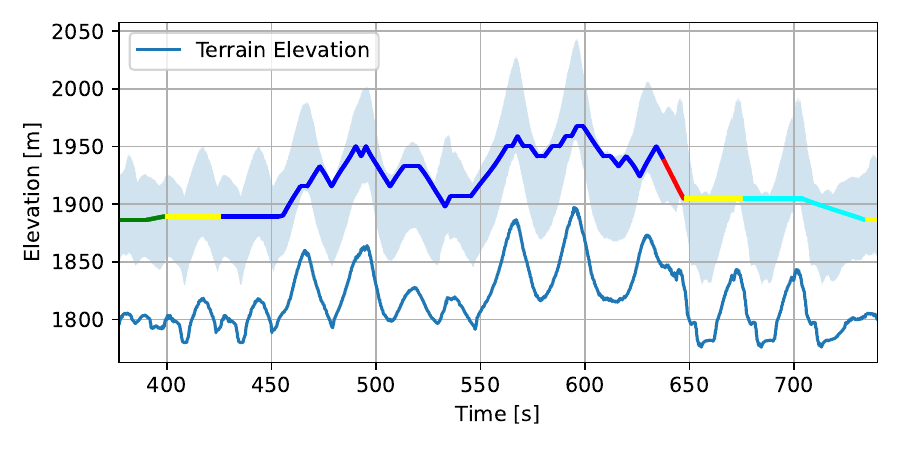}}
\caption{Altitude of the reference position and terrain visualized during the mission. The reference is color-coded with the vehicle state as in~\reffig{fig:fsm}. The vehicle satisfies the distance-to-terrain constraints visualized as the blue overlay.}
\label{fig:davos_reference_altitude}
\end{figure}

\reffig{fig:davos_reference_altitude} visualizes the reference from the active mapping approach against time. The terrain altitude solid blue line, and the altitude in order to stay within \SI{50}{} to \SI{120}{\metre} distance to the terrain is displayed as the blue overlay. It can be seen that the planned path does not violate the distance constraints throughout the whole mission. Note that the steep regions of the state space only permits a narrow corridor for the vehicle to cross due to the limited climb rates. 

\begin{figure}[t]
\centerline{\includegraphics[width=\columnwidth]{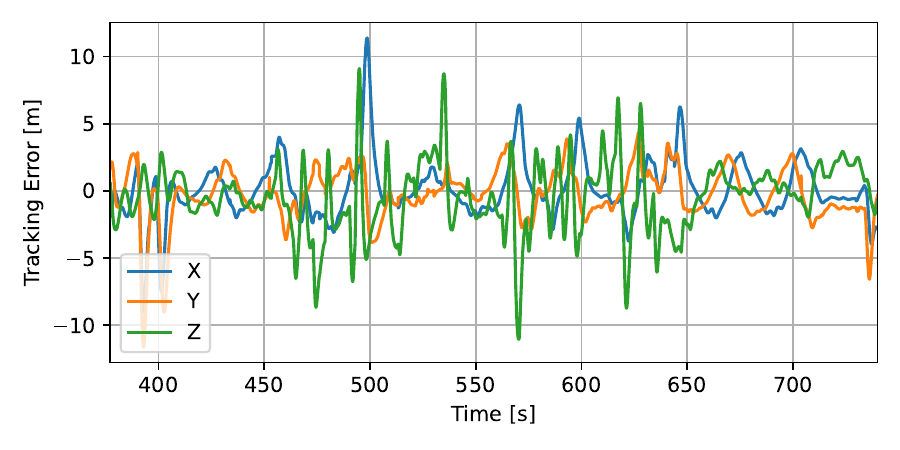}}
\caption{Tracking error of the vehicle to the reference. The large tracking errors come from the discontinuous changes in curvature and flight path angle of the reference path.}
\label{fig:davos_tracking_error}
\end{figure}

\reffig{fig:davos_tracking_error} visualizes the axis-wise tracking error throughout the mission. The large tracking errors are most significant when there are large discontinuous curvature changes or flight path angles in the path. This also means that while the planned references are under the terrain distance constraints, the vehicle may not be within the distance constraints.

\subsection{Active Mapping}

Viewpoints and vehicle paths generated from the active mapping method is shown in \reffig{fig:hinwil_mapping_results:a} and coverage planning in \reffig{fig:hinwil_mapping_results:b}. The coverage planning method is based on generating sweep patterns from a specified sweep direction and paths between the sweep patterns are planned considering the terrain distance constraints. Qualitatively, the proposed active mapping approach continuously circles the region of interest, while the coverage planning approach follows a sweep pattern.

\reffig{fig:hinwil_mapping_results:c} shows the expected reconstruction uncertainty and coverage during the mission. The expected reconstruction uncertainty is the average value of the \CR bound as described in~\cite{lim_fisher_2023}. The coverage is the proportion of the map that was visible from more than two viewpoints. We can see that the active mapping approach can generate viewpoints that achieve lower uncertainty while achieving higher coverage within a given time. This is because the active mapping approach takes advantage of the oblique views to achieve higher coverage and lower uncertainty. Also, the coverage planning method doesn't achieve full coverage, as some viewpoints were not pointing directly towards the nadir direction due to the wind present during mapping.

\begin{figure}[t]
\vspace*{0.2cm}
\begin{subfigure}{0.49\linewidth}
\centerline{\includegraphics[width=\linewidth]{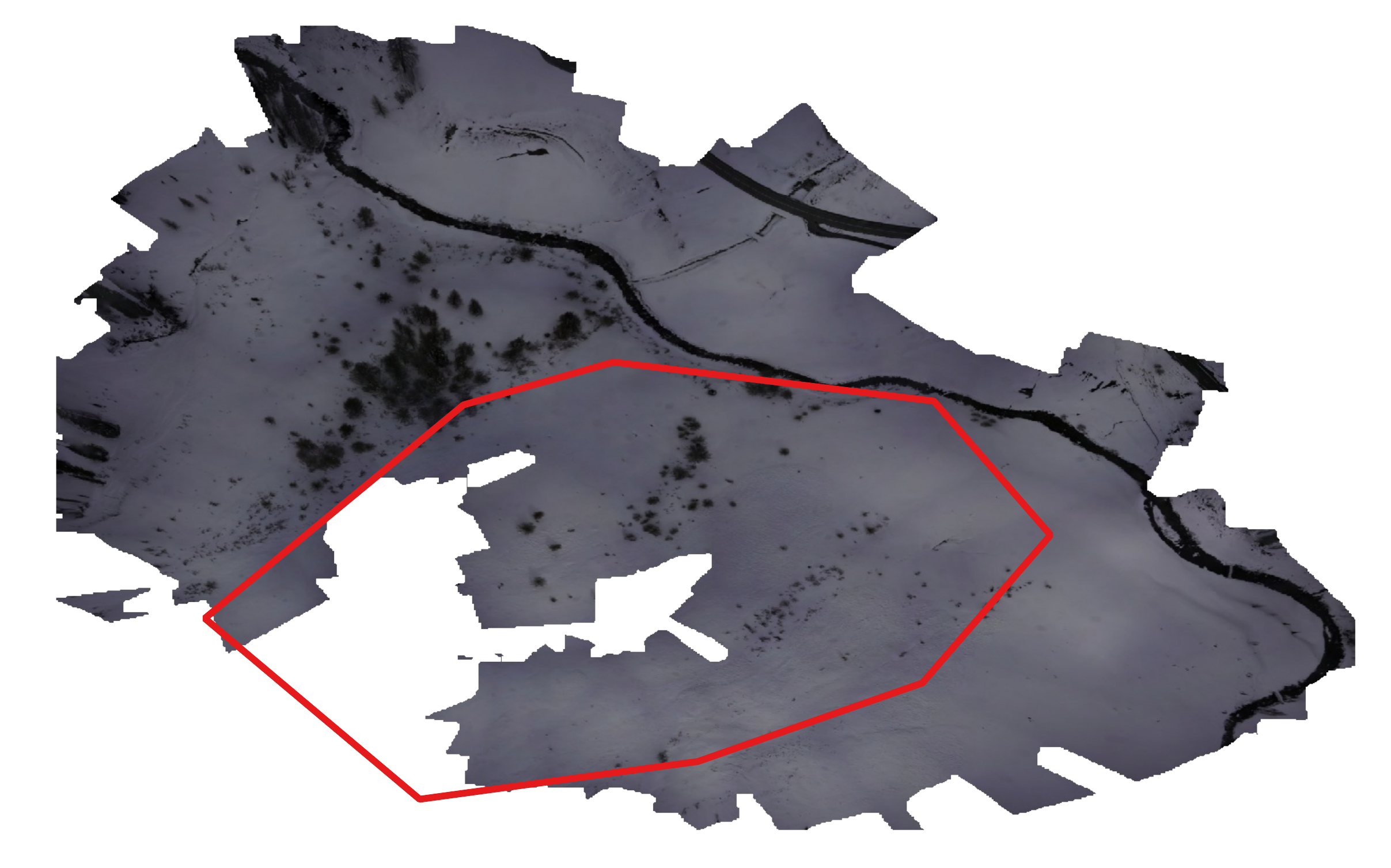}}
\caption{Active Mapping.}
\label{fig:hinwil_mapping_qualitative:a}
\end{subfigure}
\begin{subfigure}{0.49\linewidth}
\centerline{\includegraphics[width=\linewidth]{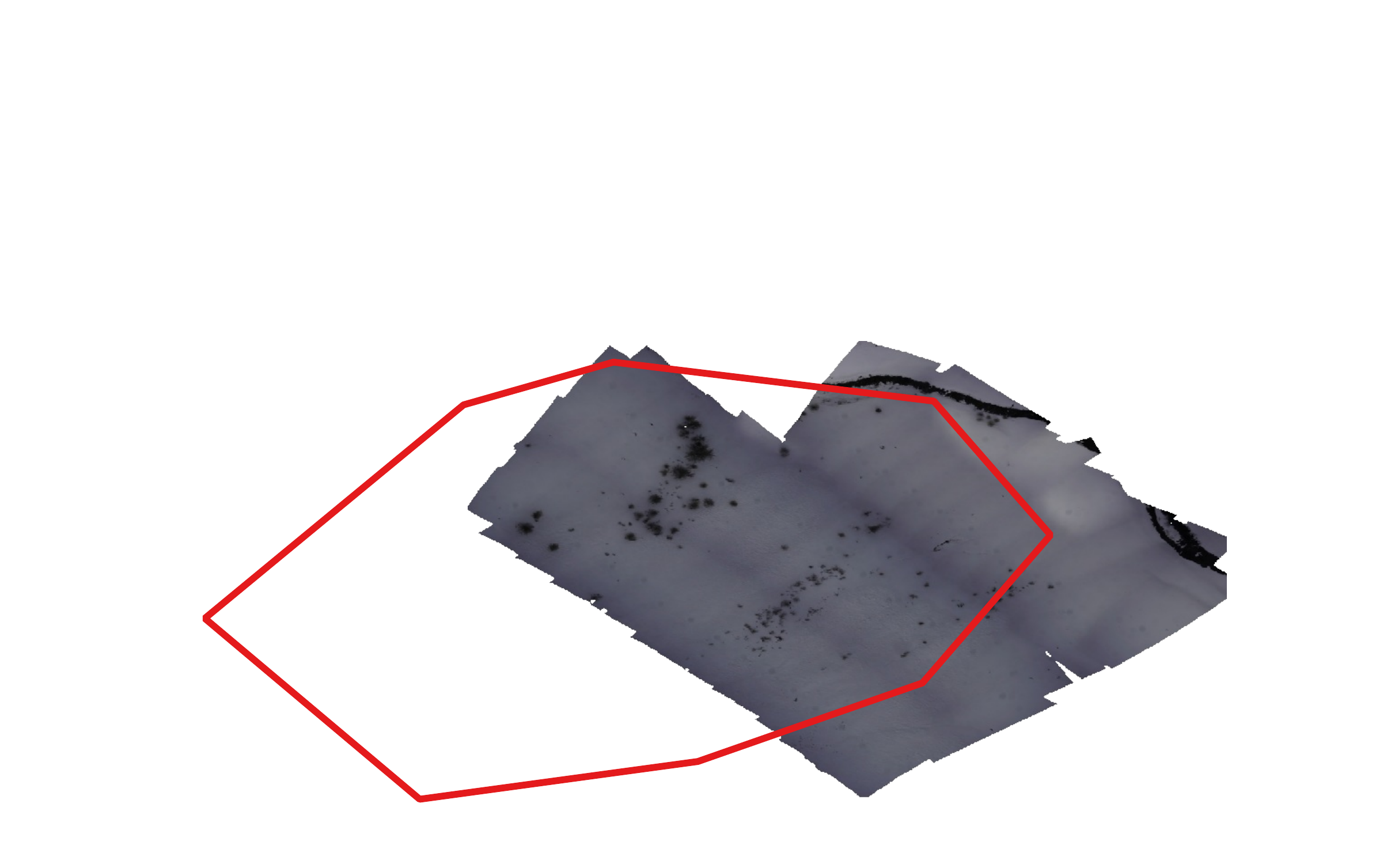}}
\caption{Coverage mapping.}
\label{fig:hinwil_mapping_qualitative:b}
\end{subfigure}
\caption{Qualitative comparison of orthomosaic reconstruction. The red polygon represents the region of interest.}
\label{fig:hinwil_mapping_qualitative}
\end{figure}

\reffig{fig:hinwil_mapping_qualitative} shows a qualitative comparison of the orthomosaic reconstruction results from Agisoft. Note that the predicted uncertainty metric in~\reffig{fig:hinwil_mapping_results:c} predicted the reconstruction to be complete, there is a large hole within the region of interest in~\reffig{fig:hinwil_mapping_qualitative:a}. This is a region where the feature density was low due to the smooth snow surface. The planner is susceptible to such scenes as the view utility metric in~\cite{lim_fisher_2023} only considering the camera network geometry without the consideration of photometric appearance. This would introduce discrepancies with the information estimate on the surface where the feature density is low. The effect of low texture seems to be more severe for the coverage mapping as shown in~\reffig{fig:hinwil_mapping_qualitative:b}, since the surface is always viewed from the same direction, making it more susceptible to bad feature matches.

\section{DISCUSSIONS}
We have demonstrated an integrated fixed-wing system capable of safely operating close to terrain and successfully executing an information-gathering task. We show that the limited maneuverability of fixed-wing vehicles can be effectively handled by planning kinematically constrained paths and periodic goal sets. We also demonstrate an information-gathering task that can be deployed with the high operating speed of fixed-wing vehicles, by actively optimizing viewpoints for aerial photogrammetry. However, challenges remain in making such systems reliably operate autonomously in real-world operations.

The integration of finite state machine and safe planner provides a simple interface to operate the vehicle through high-level commands. This simplifies the operational complexity of fixed-wing and hybrid \ac{VTOL} vehicles. The flight tests show large tracking errors due to the discontinuous curvature changes and flight path angle changes. Such limitations could be overcome with more sophisticated path representations or model predictive controllers which would better consider the maneuverability constraints. However, the sensitivity of the vehicle dynamics to wind, and the large uncertainty associated with wind fields make it challenging to use more complicated path representations. Works to predict wind fields have been considered~\cite{achermann_windseer_2024}, but the spatio-temporal nature of winds and the complex interaction with the environments make handling wind a major challenge. Being able to reliably predict wind fields will not only improve safety but also provide opportunities to exploit such wind fields~\cite{duan_energyoptimized_2024, lawrance_autonomous_2011}.

One of the main limitations of the approach is that the methods use a strong prior provided by \acp{DEM} from a~\ac{GIS}. While such priors are useful for planning, natural environments are always evolving. Therefore, perception capabilities to to safely react to unknown environmental factors are crucial for operating in large-scale natural environments. Due to the high operating speed of fixed-wing vehicles, sensors with sufficiently long ranges such that the vehicle can take evading action are needed. However, the availability of sensors for this application is currently limited. Moreover, technical and regulatory challenges for sense-and-avoid systems remain open.

The active view planning approach has demonstrated that an active view planner can generate viewpoints that result in more accurate reconstruction results for photogrammetry. However, field tests have shown that a purely camera network geometry based approach as used in~\cite{lim_fisher_2023} is not sufficient to ensure good reconstruction results in challenging scenes such as snow. Methods considering appearance in addition to the camera network geometry~\cite{liu_learning_2022} could be useful, but processing high-resolution image data in real-time might be challenging. Lastly, the active view planning approach plans maneuvers in a receding horizon manner. This makes the maneuver susceptible to encountering \iac{ICS}, as large parts of the environment are too steep to fit a loiter path as shown in~\reffig{fig:terrain_safetymask}. Therefore, the problem of ensuring safety while incrementally planning maneuvers needs to be addressed.

\section{CONCLUSIONS}

In this workshop paper, we have looked into the challenges of enabling environment monitoring tasks in steep alpine terrain with a fixed-wing aerial vehicle. We have demonstrated a fixed-wing \ac{sUAS} capable of safely navigating steep terrain and actively mapping a target region of interest in a field test. We have shared our preliminary results from the field test and showed that the mapping performance outperforms the results when planned with a coverage path. 
We believe autonomous fixed-wing aerial vehicles will be a powerful tool for gathering high-quality data for large-scale environment monitoring applications.









\bibliographystyle{IEEEtran}
\bibliography{references}

\end{document}